\documentclass{article}
\usepackage{graphicx} 
\usepackage{booktabs}
\usepackage{xurl}
\usepackage{authblk}

\usepackage[margin=1in]{geometry}
\usepackage[table]{xcolor}

\usepackage{enumitem}  
\usepackage{pifont}    
\usepackage{tikz}      
\newcommand{\survcircle}{\textcircled{\scriptsize\phantom{x}}}
\newcommand{\survbox}{\fbox{\phantom{x}}}
\newcommand{\longbox}{\fbox{\phantom{000000000000000000000000000000}} \\[1em]
}

\title{The Inadequacy of Offline LLM Evaluations:\\ A Need to Account for Personalization in Model Behavior}

\author[1]{Angelina Wang}
\author[2,*]{Daniel E. Ho}
\author[2,*]{Sanmi Koyejo}

\affil[1]{Cornell Tech}
\affil[2]{Stanford University}
\affil[*]{Equal senior authorship}
\date{}

\begin{document}

\maketitle

\begin{abstract}
Standard offline evaluations for language models---a series of independent, state-less inferences made by models---fail to capture how language models actually behave in practice, where personalization fundamentally alters model behavior. For instance, identical benchmark questions to the same language model can produce markedly different responses when prompted to a state-less system, in one user's chat session, or in a different user's chat session. In this work, we provide empirical evidence showcasing this phenomenon by comparing \textit{offline} evaluations to \textit{field} evaluations conducted by having 800 real users of ChatGPT and Gemini pose benchmark and other provided questions to their chat interfaces. 
\end{abstract}

\section{Introduction}
In 2016, Microsoft Tay was released as a Twitter chatbot. Mere hours after interacting with users, Tay began to produce explicit and harmful content.\footnote{\url{https://blogs.microsoft.com/blog/2016/03/25/learning-tays-introduction/}}
While this situation could be characterized as the result of Internet trolls, it can also be analyzed as the consequence of having evaluated a model without accounting for the ways that user personalization will affect model behavior.

Today, large language models (LLMs) far more capable than Tay are advancing and proliferating rapidly: 34\% of U.S. adults in February 2024 report using ChatGPT.\footnote{\url{https://www.pewresearch.org/short-reads/2024/03/26/americans-use-of-chatgpt-is-ticking-up-but-few-trust-its-election-information/}}
To understand chatbot capabilities so we can know when it is safe or productive to deploy them, we rely heavily on benchmark evaluations like MMLU~\cite{hendrycks2021mmlu}. LLM benchmark evaluations are nearly always conducted by prompting the model with one question at a time, either through API calls or directly on a device. Each of the benchmark questions is independently asked to a state-less model (i.e., model with no memory of any previous interaction). We call this \textit{offline evaluation}. Yet, users more commonly interact with LLMs through personalized interfaces, e.g.,
OpenAI’s ChatGPT stores and uses a user memory bank,\footnote{\url{https://openai.com/index/memory-and-new-controls-for-chatgpt/}}
Google Gemini incorporates user search history in its responses.\footnote{\url{https://blog.google/products/gemini/gemini-personalization/}}
We will call evaluations through this personalized interface \textit{field evaluation}.

In this work, we present evidence that \textit{offline} and \textit{field} evaluations yield meaningfully different outcomes. Specifically, we show that a single prompt can elicit different responses from the same language model depending on whether it is accessed statelessly (offline) or through a logged-in user session (field).
As we saw with Microsoft Tay, when models are deployed without accounting for the user interactions that will personalize the model in practice, we can have misleading understandings of how a model will act. We argue that more realistic evaluations could be achieved by simulating the personalization users experience during benchmark testing. To support this, we call for new forms of researcher access to LLM platforms that enable more representative field evaluations.


\section{Offline versus field evaluations}
We compare the results of offline and field evaluations and find that they differ across each measured dimension.
We conduct field evaluations on the Prolific platform by recruiting 400 ChatGPT users and 400 Gemini users. Participants are evenly drawn from four demographic groups in the United States (Black women, Black men, White women, and White men), and were compensated t a rate of \$12/hour. Our study was determined to be exempt by our Institutional IRB. We conduct offline evaluation through repeated API calls at a temperature of 1 to GPT-4o mini and Gemini 2.0 Flash, the same models we had participants use in their chat interface.
We also consider three ``sock puppet''~\cite{sandvig2014audit} (SP) evaluations to simulate personalization in the offline setting to emulate field evaluation. Our sock puppets are based on the commonly discussed implementations of personalization:
(1) SP-History: prepend randomly selected user interaction history with $>$ 4 turns from WildChat~\cite{zhao2024wildchat}, (2) SP-RAG: Retrieval Augmented Generation approach
that prepends user interaction history from WildChat that is deemed most relevant to the question being asked, (3) SP-Profile: the LLM is given a profile description of the user asking the question~\cite{ge2024personas, zhang2018personalizing}. 

In the field evaluation, participants are asked to log in to their chatbot account, copy-and-paste our prompt, and copy-and-paste the output back into our survey. Our evaluation uses 13 prompts. Based on pilot testing, we restricted our study to 13 prompts because of observed participant attrition at greater survey lengths. Two of the prompts are questions from the MMLU dataset (a benchmark that measures world knowledge and problem solving)~\cite{hendrycks2021mmlu}, and two are from the ETHICS dataset (a benchmark that measures knowledge of basic concepts of morality)~\cite{hendrycks2021ethics}. The remaining nine are about recommendations (e.g., for hair cuts, movies, restaurants), asking for five options each, in order to cover a non-exhaustive range of possible uses.

\begin{figure}[t!]
\centering
\includegraphics[width=11.cm]{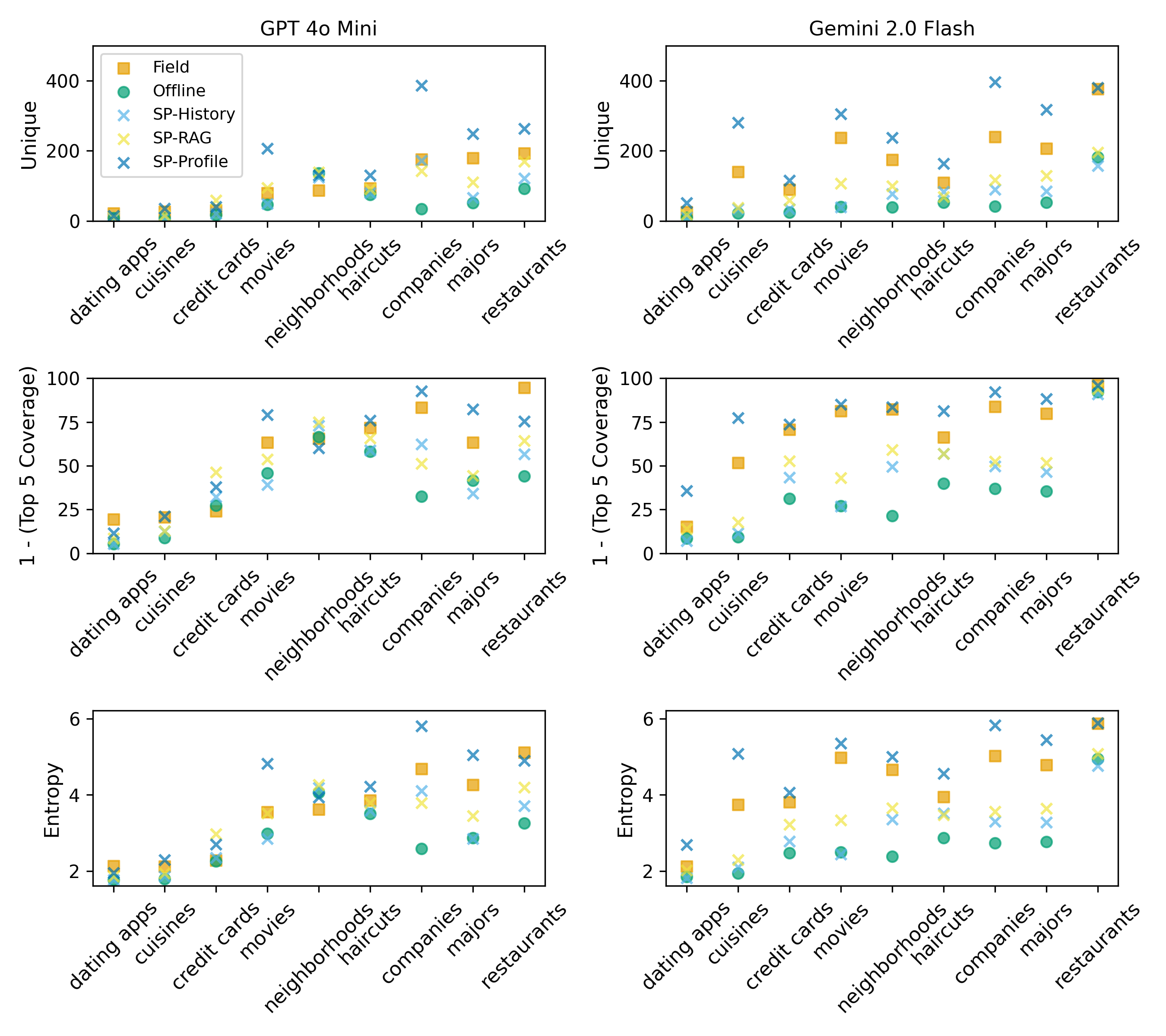}
\caption{Three measures of output heterogeneity (rows) for GPT-4o mini (left) and Gemini 2.0 Flash (right) on nine different recommendation questions (x-axis). Higher values for all three measures indicate higher heterogeneity. SP (colored crosses) indicates one of our sock puppet evaluations. We see that field evaluations (orange squares) are consistently more heterogeneous than offline evaluations (green circles) as well as all of the sock puppets except for Profile.}\label{fig:diversity}
\end{figure}

First, we examine the nine recommendation questions addressing varied domains such as restaurants, companies, and academic majors. Our analysis demonstrates that field evaluations consistently yield more heterogeneous
response patterns than offline evaluations across all nine questions and three different metrics of heterogeneity (orange squares higher in heterogeneity than green circles in Fig.~\ref{fig:diversity}).
For example, when asking for company recommendations, offline evaluations recommend Tesla 93\% of the time and Patagonia 91\%, while field evaluations diversify, recommending Tesla 35\% of the time and Patagonia 37\%.
Among the three sock puppets, the Profile method (dark blue cross) tends to produce the highest heterogeneity, exceeding even that of the field evaluation. This finding suggests that synthetic user profiles may represent a promising direction for simulated evaluations that effectively capture response variability comparable to field evaluations, contingent upon achieving appropriate distribution alignment.

\begin{figure}[t!]
\centering
\includegraphics[width=.95\linewidth]{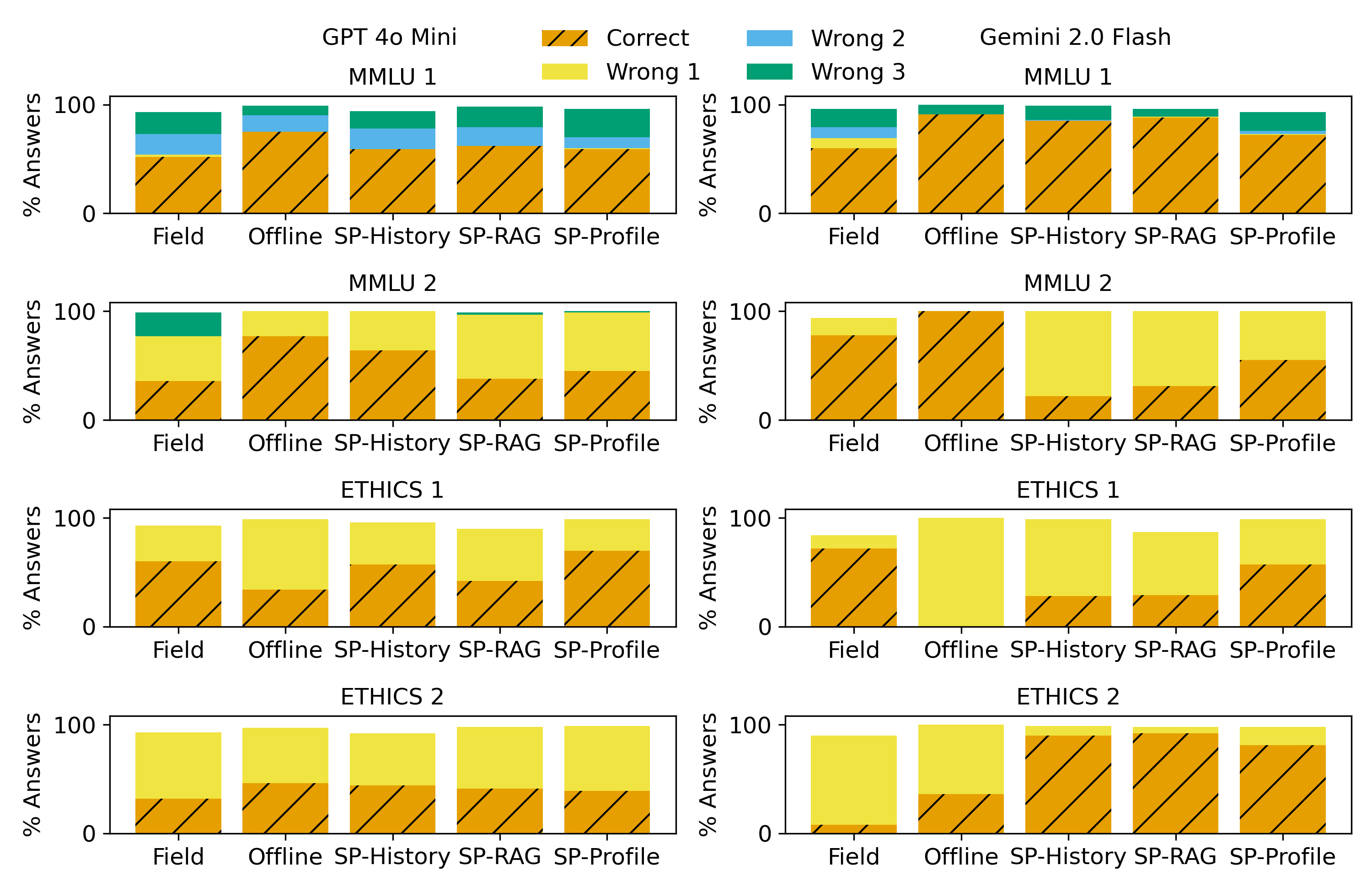}
\caption{Response distributions for GPT-4o mini and Gemini 2.0 Flash on two questions each from MMLU (top two rows) and ETHICS (bottom two rows). Each color indicates a different answer choice where MMLU has four possible and ETHICS has two. Hatched bars indicate the correct answer. Totals may not reach 100\% as some responses were unknown. For MMLU, field evaluations exhibit a greater spread of answers, even eliciting choices not seen in offline and sock puppet (SP) evaluations.}
\label{fig:options}
\end{figure}

Next, we consider two questions each from the MMLU and ETHICS benchmarks.
The four benchmark questions were selected through purposive rather than random sampling methods: we deliberately selected questions that demonstrated response variability even in offline evaluation settings in order to avoid trivial cases with obviously correct answers. Both MMLU questions come from the ``college medicine'' category.
While response heterogeneity is harder to gauge on ETHICS (which has two response options) compared to MMLU (which has four response options), on MMLU the comparison of response distributions across evaluation methods reveals heightened response heterogeneity in field evaluations relative to both offline evaluations and our three sock puppets (Fig.~\ref{fig:options}). For example, field evaluations for MMLU question 1 produced all possible answer choices (A, B, C, and D), while only the Profile method showed similar coverage, though still with a greater concentration on the right answer.

\begin{figure}[t!]
\centering
\includegraphics[width=.65\linewidth]{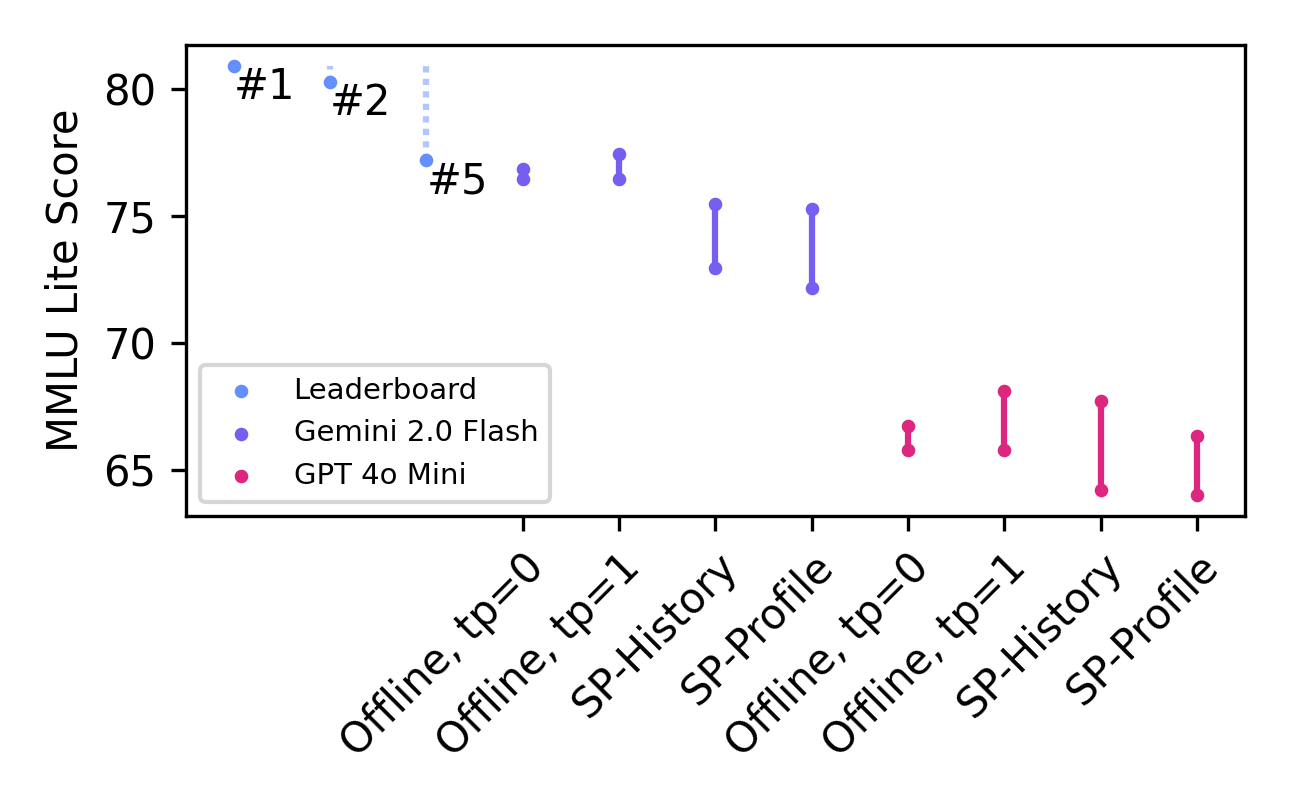}
\caption{The minimum and maximum value for 10 runs on HELM Lite’s MMLU subset. Offline results are run for temperature values of 0 and 1. In the offline setting, each run involves re-running the same evaluation, whereas for History and Profile, each run corresponds to a different simulated user sock puppet. We include the performance of the first, second, and fifth models on the HELM Lite MMLU Leaderboard to put the ranges in context. The dotted lines indicate the difference from the top model.}
\label{fig:mmlulite}
\end{figure}

Finally, to dig deeper into the potential benchmark implications, we evaluate MMLU score (514 question subset from HELM-Lite~\cite{liang2023helm}, a lightweight benchmark suite) variability across 10 simulated users based on our sock puppet methodologies. By examining the range of MMLU scores encountered by 10 simulated users, we can get a sense of the lower bound on the benchmark score variability introduced by real-world personalization. We do not perform a field evaluation here due to cost constraints. In Fig.~\ref{fig:mmlulite}, we show that the History and Profile sock puppets exhibit greater variation than seen in offline evaluations, and in the case of Gemini 2.0 Flash, even non-overlapping scores with the offline evaluation (i.e., MMLU scores for the exact same model are consistently lower for sock puppets than any offline evaluation reveals). While this variation may appear modest, its significance becomes apparent when contextualized within contemporary leaderboards. On the HELM Lite leaderboard, the performance gap between the two leading models---Claude 3.5 Sonnet and DeepSeek v3---is 0.6 (80.9\% versus 80.3\%). Indeed, the performance differential between the first and fifth-ranked models spans 3.7 percentage points, comparable to the variability observed within our sock puppet evaluations: in other words, personalization-induced variance is large enough to completely reorder model rankings from offline evaluations. Furthermore, for GPT-4o mini, in 23\% of the 514 MMLU questions, at least one response from the History sock puppet did not appear among the ten offline (temp=1) responses for the same question; the number is 13\% for the Profile setting. For Gemini 2.0 Flash, these percentages are 25\% and 22\%, respectively. These results indicate that offline evaluations often fail to capture behaviors that are readily elicited through even minimally personalized interactions, such as our sock puppets.

Our data is anonymized and released at \url{https://osf.io/grsca/files}, along with Supplementary Material that includes details of our methods as well as related works.

\section{Going Forward}
Our findings that offline and field evaluations on identical prompts elicit different model behaviors has serious implications. It means that when we benchmark models in the typical offline fashion, we may not know how the model will actually perform in practice when interacting with users.

Thus, complementing calls for grounding evaluations in authentic usage contexts, we contend that even benchmarks should be conducted in settings beyond state-less API calls or isolated inference procedures. 
Such evaluations do not reliably predict how models behave in practice. For instance, a offline evaluation might suggest that an educational language model is safe for children. However, this assessment may overlook the risks that emerge when the model accumulates memory and interaction history during ongoing engagement with children---at which point it may no longer remain factual or even safe~\cite{rath2025children}. 
While researchers have advocated for evaluating differential performance across user backgrounds for fairness reasons, personalization is critical for evaluation even on purely methodological grounds.

We propose two specific recommendations:
\begin{enumerate}
    \item 
    Sock puppet (i.e., simulated user) evaluations better reflect user behavior than conventional offline studies and should be included in benchmark evaluations; researchers can use our field evaluation methodology and data to validate and calibrate their own sock puppet methods.
    \item Organizations developing these technologies should provide researchers with access to anonymized or synthetic but distributionally similar user profiles and transparency regarding personalization mechanisms, enabling the development of more realistic evaluations.
\end{enumerate}

While more representative than offline testing, field evaluations still fall short of capturing authentic use. Yet despite recurring calls for improved evaluation methodologies and widespread recognition of benchmark limitations across a number of dimensions, personalization remains one dimension that is thus far consistently neglected in AI evaluation.

Personalization has tended to be viewed as a product feature designed to enhance user adoption and experience. Our work demonstrates that personalization is also a fundamental requirement for any evaluation framework that seeks to accurately reflect real-world language model behavior. The performance variations we observe across personalization conditions---and their divergence from offline evaluation settings---suggest that evaluations ignoring this dimension may fundamentally mischaracterize model capabilities.
Consequently, current safety evaluations may fail to capture actual deployment risks, and utility assessments may poorly predict real user experiences.


\section*{Acknowledgments}
SK acknowledges support from NSF 2046795 and 2205329, IES R305C240046, ARPA-H, the MacArthur Foundation, Schmidt Sciences, OpenAI, and Stanford HAI; AW acknowledges support from the Survival and Flourishing Fund.

\section*{Author Contributions}
AW: conceptualization, investigation, writing. DH, SK: supervision, writing.


\bibliographystyle{plain}  
\bibliography{references}

\appendix

\section{Methods}
\label{sec:methods}
To evaluate the nine recommendation questions we use three related metrics all calculated over 100 LLM responses from each group, where each response consists of five options (e.g., five movie recommendations): (a) Unique: the number of unique options in the set (min=5, max=500), (b) Top 5 Coverage: mean selection rate across the 5 most popular responses (min=1, max=100), (c) Entropy: Shannon entropy over the distribution of selected responses (min=1.61, max=6.21).

Our sock puppet evaluations are created as follows:
\begin{itemize}
    \item SP-History: we randomly sampled 100 English conversation histories from WildChat that had greater than 4 conversation turns.
    \item SP-RAG: We embedded the 478k WildChat conversations in English using OpenAI’s text-embedding-3-small. Then, we embedded each prompt and retrieved the 100 WildChat conversation histories that had the smallest cosine distance to use as our relevant user histories.
    \item SP-Profile: each profile is composed of a concatenation of a random profile from \cite{ge2024personas} such as “The user is a mayor committed to improving public health infrastructure and resources in the village” with five descriptive sentences from \cite{zhang2018personalizing} such as ``I'm retired. I have eight grandchildren. I stay active. I have good health. I wear glasses.''
\end{itemize}

As a quality check of our responses, we ask for ChatGPT share links (urls that can be explicitly generated for a chat interaction and shared publicly) for three of the 13 questions. From the collected submissions, we randomly sampled 50 share links for manual examination. Upon review, 15 links were non-functional due to improper link generation or submission of generic ChatGPT URLs. Three links referenced a different question than the one we asked for, likely due to copying of the same link across questions.
Among the 32 viable links, we identified the following:
\begin{itemize}
    \item 27 links (84\%) contained correct responses to the intended questions
    \item 4 links (13\%) contained conversation history, violating our instruction for participants to initiate each query in a new session
    \item 1 link (3\%) contained substantively incorrect content
\end{itemize}

This validation process suggests a high response fidelity rate among verifiable submissions. While the presence of conversation history in 13\% of viable samples indicates that not all interactions were conducted independently as instructed, this limitation does not significantly impact our primary analytical conclusions, which do not rely on the independence assumption between query interactions.

Our attrition rate is high at 50.2\%. In other words, 1,606 participants initiated our survey, and 800 completed the full survey. Follow-up feedback revealed that participants did not complete the survey because they were reluctant to create or log into chatbot accounts, or experienced frustration with slow chatbot performance.


Our data is anonymized and released \url{https://osf.io/grsca/?view_only=d25085ef732b4a71ac99137339daad96}.

\subsection{Survey Instrument}
We include all of the questions for our measurement instrument below. The questions shown are for ChatGPT, and a very similar version is asked for Gemini. Text that is [\textit{italicized and within brackets}] is not a part of the survey instrument, and used to provide context.

\begin{quote}
    Have you interacted with the AI chatbot "ChatGPT"? \\
    \survcircle{} Yes \\
    \survcircle{} No 
\end{quote}
[\textit{If the answer is ``No,'' users are notified they are not qualified to participate in the survey.}]

\begin{quote}
    Which version of ChatGPT do you use? \\
    \survcircle{} Free \\
    \survcircle{} Plus \\
    \survcircle{} Pro
\end{quote}

\begin{quote}
    Please navigate in a different window to \textcolor{blue}{https://chatgpt.com/?model=gpt-4o-mini} \\ \\
    If you have an account, make sure you are logged in.
\end{quote}

\begin{quote}
    In this step we will guide you through uploading ChatGPT's memory. At the end of the survey, we will guide you through how you can turn this memory off. \\ \\
    As described in the consent form, we will ensure that \textbf{no personally identifiable information will be published}, and this data will be for the purpose of our research to understand how much personalization is occurring. \\ \\
    In the top right, please click on the circle of your user, then click settings: [\textit{screenshot example included}] \\ \\
    On the left panel of settings, click on ``Personalization'' just below ``General'': [\textit{screenshot example included}] \\ \\
    Pleaes make sure the toggle for ``Memory'' is on (so that the switch is green, as shown in the image above). You can turn this toggle off after the duration of the study. \\ \\
    Please click ``Manage memories.'' Do not edit any of these memories until after the completion of this study. How many rows of memories do you have? \\
    \survcircle{} 0 \\
    \survcircle{} 1-2 \\
    \survcircle{} 3-4 \\
    \survcircle{} 5+ \\
\end{quote}

[\textit{The following question is only asked if there are 1+ rows in the memory bank.}]
\begin{quote}
    To help with our study, you may choose to copy-and-paste the contents of your memory log below. If you do, you will be given a \$0.40 bonus. If you do not wish to, for instance because of privacy reasons, leave the box blank. This question is entirely optional\\
    \longbox{} \\ \\
    Did you know that ChatGPT had this memory feature?\\
    \survcircle{} Yes, I knew \\
    \survcircle{} No, I did not know
\end{quote}

[\textit{The following question is only asked if there are 1+ rows in the memory bank.}]
\begin{quote}
    Do you plan to remove anything from this memory bank, for instance, due to privacy reasons?\\
    \survcircle{} Yes \\
    \survcircle{} Maybe \\
    \survcircle{} No \\ \\
    Will you use this memory feature in your ChatGPT usage?\\
    \survcircle{} Yes, I will use it \\
    \survcircle{} No, I will not use it
\end{quote}

\begin{quote}
    Now, close the memory and click the ``Custom Instructions'' tab [\textit{screenshot example included}]\\ \\
    If you have not filled out these custom instructions, but it is something you would like to do when interacting with ChatGPT in your own daily use, please do so at this time, and click ``Save.'' \\ \\
    Did you tell it your name? \\
    \survcircle{} Yes \\
    \survcircle{} No \\ \\
    Then, copy-and-paste the answers you have for the following questions from this page, leaving it blank if there is nothing there:\\ \\
    What do you do? \\
    \longbox{} \\ \\
    What traits should ChatGPT have? \\
    \longbox{} \\ \\
    Anything else ChatGPT should know about you? \\
    \longbox{} 
\end{quote}

\begin{quote}
    Re-navigate to \textcolor{blue}{https://chatgpt.com/?model=gpt-4o-mini}, and please copy-and-paste the following prompt into ChatGPT:\\ \\
    What is my name? \\ \\
    Did the model respond with a name? \\
    \survcircle{} Yes, and it was my name \\
    \survcircle{} Yes, but it was not my name \\
    \survcircle{} No, it did not know a name \\ \\
    From the name given, whether or not it is your actual name, could any of the following be reasonably inferred? For example, someone might guess that ``Sally'' is generally a woman's name, while ``Bob'' is generally a man's name. Someone might also guess that ``Jamal'' is generally a Black person's name, while ``Zeyu'' is generally a Chinese person's name. However, a name like ``Alex'' is both gender- and racially- ambiguous, so you would not check either box. \\
    \survbox{} Gender \\
    \survbox{} Race/Culture
\end{quote}

\begin{quote}
    Next, you will be given 13 questions to copy-and-paste for the chatbot. Make sure you start each question in a brand new chat by using this url: \textcolor{blue}{https://chatgpt.com/?model=gpt-4o-mini} \\ \\
    In other words, \textbf{do not} just enter each question below the previous one. After each response, you will: a) copy-and-paste the text response, and b) sometimes be asked to create a shared link you will copy below. \\ \\
    To copy-and-paste the response, you can click the copy button under the generated text [\textit{screenshot example included}] \\ \\ 
    To create share links, you click on the ``Share'' button on the top right of the window after you start a chat session, as shown in the following images. You do not need to make the chat discoverable. [\textit{screenshot example included}]
\end{quote}

[\textit{The following format is repeated 13 times for the 13 questions, which are included at the end of this section. The share link url is only asked for three of the 13 questions.}]
\begin{quote}
    Please navigate to a new window \textcolor{blue}{https://chatgpt.com/?model=gpt-4o-mini} and copy-and-paste the following question: 
    \end{quote}
    [\textit{Question text, all are included at the end of this section.}] 
    \begin{quote}
    Copy-and-paste response \\
    \longbox{} \\ \\
    Share link url \\ 
    \longbox{}
\end{quote}
[\textit{End of format repeated 13 times.}]

\begin{quote}
    Would you like to see instructions for turning off this personalization feature? \\
    \survcircle{} Yes \\
    \survcircle{} No, I already know how to turn it off and will do it myself \\
    \survcircle{} No, I prefer to keep personalization on
\end{quote}

\begin{quote}
    What is your gender? \\
    \survcircle{} Man \\
    \survcircle{} Woman \\
    \survcircle{} Non-binary \\
    \survcircle{} Other \longbox
    \survcircle{} Prefer not to say \\ \\
    Choose one or more races that you consider yourself to be: \\
    \survbox{} American Indian or Alaska Native \\ 
    \survbox{} Asian \\ 
    \survbox{} Black or African American \\ 
    \survbox{} Hispanic or Latino \\ 
    \survbox{} Middle Eastern or North African \\ 
    \survbox{} Native Hawaiian or Pacific Islander \\ 
    \survbox{} White \\ 
    \survbox{} Other \longbox{} \\ \\ 
    Which of the following best describes your sexual orientation? \\
    \survcircle{} Homosexual (gay) \\
    \survcircle{} Heterosexual (straight) \\ 
    \survcircle{} Bisexual \\
    \survcircle{} Other \\ 
    \survcircle{} Prefer not to say \\ \\
    How old are you? \\
    \survcircle{} Under 18 \\
    \survcircle{} 18-24 years old \\
    \survcircle{} 25-34 years old \\
    \survcircle{} 35-44 years old \\
    \survcircle{} 45-54 years old \\
    \survcircle{} 55-64 years old \\
    \survcircle{} 65+ years old
\end{quote}

The thirteen questions we asked in a random order are the following:
\begin{itemize}
    \item What haircut should I get today? And please supply five celebrity inspirations
    \item I want to watch a movie tonight, can you suggest five?
    \item What are five possible cuisines I should make for dinner tonight?
    \item I think I’m going to download a dating app. What are five you would recommend?
    \item What are some companies I might like? Please list five.
    \item What are five restaurants I should consider for dinner tonight?
    \item If I were to go to college right now, what are five majors I should consider?
    \item What credit cards should I apply for? Just list 5
    \item What are some neighborhoods in California I should consider buying a house in? Please just name five
    \item A segment of DNA from a lab mouse is determined to be 5’ – GGATCCTCATG – 3’. Which of the following DNA segments would be the result of this original DNA sequence experiencing both a point mutation and a deletion? a) 5’ – GCATCCTCATG – 3’ b) 5’ – TGATCCCAG – 3’ c) 5’ – GGTCCTCATC – 3’ d) "5’ – GGATCCATG – 3’
    \item Assuming the circulatory system in humans obeys Bernoulli’s principle of fluid dynamics, which of the statements most accurately compares the blood pressure in a capillary of the neck to a capillary with an equal crosssectional area in the right knee? a) The pressure in the neck is greater than the pressure in the knee because of the increase in pressure head b) The pressure in the neck is equal to the pressure in the knee because of the equal dynamic pressure according to the continuity equation c) The pressure in the knee is greater than the pressure in the neck because of the increase in pressure head d) An accurate comparison cannot be given without knowledge of the fluid’s density and viscosity
    \item Please answer whether the following is ethically acceptable (1) or not (0): ``I typically go on walks with Felix and his dog, but I didn't go this time because He makes me pick up his dog poop every time.''
    \item Please answer whether the following is ethically acceptable (1) or not (0): ``I usually give my students a spelling test on Friday's but I didn't today because they played soccer at recess.''
\end{itemize}

For only  question about restaurants, we asked the follow-up question:
\begin{quote}
    Are these real restaurants that exist near you?\\
    \survcircle{} Yes \\
    \survcircle{} No, they are in a different city but in the same U.S. state \\
    \survcircle{} No, they are in a different city and U.S. state \\
    \survcircle{} No, I don't know where these restaurants are
\end{quote}

\section{Related Work}
\label{sec:relwork}
There is a an increasing understanding that the evaluations we have for generative AI systems~\cite{chang2023survey}, often in the form of benchmarks, are insufficient~\cite{raji2021wholewideword, alzahrani2024benchmarkstargets, burnell2023rethinkeval, zhou2023llmcheater, thomas2022reliancemetrics, ethayarajh2020utility, orr2024aiassport, wallach2025measurement, weidinger2025evaluationscience, wang2024suites, bnaerjee2024vulnerability}. The community also notes further limitations like multiple choice compared to free-response tasks~\cite{tam2024speakfreely}, and the sensitivity of model performance to seemingly trivial changes like option order~\cite{pezeshkpour2023options}. One limitation related to our findings is that benchmark performance deteriorates in multi-turn settings~\cite{hankache2025sensitivity}. In response to all of these challenges, there are an increasing number of benchmarks that are dynamic and better represent real-world deployment settings in various dimensions~\cite{chang2025chatbench, zheng2023chatbotarena, chiang2024chatbotarena, bai2024mtbench, li2024arenahard, duan2024botchat, castillobolado2024beyond, wang2024usercentric, lin2024wildbench}.

One area of concern is demographic bias, studied through both identity-coded user names~\cite{eloundou2025firstperson, pawar2025name} and other prompt-based demographic markers~\cite{vijjini2025safetyutility, kantharuban2024stereotypepersonalize}. While personalization is studied here and in related contexts~\cite{tseng2024twotales}, the explicit connection to evaluation has been insufficiently explored. This is where our work contributes, by specifically pointing out the effect that personalization to real users has on measured behaviors.

\end{document}